\documentclass{article}
\usepackage[utf8]{inputenc}
\usepackage{graphicx}
\usepackage[english]{babel}
\usepackage{tikz}
\usepackage{color}
\usepackage{amsthm}
\usepackage{bm}
\usepackage{amsmath}
\usepackage{amsmath,amssymb}
\usepackage{graphicx}
\usepackage[colorinlistoftodos]{todonotes}
\usepackage[colorlinks=true, allcolors=blue]{hyperref}
\usepackage[T1]{fontenc}
\usepackage[square,numbers]{natbib}

\usepackage[a4paper,
            bindingoffset=0.2in,
            left=1in,
            right=1in,
            top=1in,
            bottom=1in,
            footskip=.25in]{geometry}

\usepackage{epsfig}
\usepackage{amsmath}
\usepackage{amssymb}
\usepackage{booktabs}
\usepackage{xcolor,colortbl}
\usepackage{multirow}
\usepackage{subcaption}
\usepackage{url} 
\usepackage[normalem]{ulem}
\usepackage{tabularray}
\usepackage{float}

\usepackage[capitalize]{cleveref}
\crefname{section}{Sec.}{Secs.}
\Crefname{section}{Section}{Sections}
\Crefname{table}{Table}{Tables}
\crefname{table}{Tab.}{Tabs.}

\definecolor{mygray}{gray}{0.9}

\usepackage{algorithm}
\usepackage{algorithmic}

\title{PA-SAM: Prompt Adapter SAM for High-Quality \\Image Segmentation}
\author{Zhaozhi Xie$^{1,2}$, Bochen Guan$^{2,3}$, Weihao Jiang$^{1}$, Muyang Yi$^{1}$, Yue Ding$^{1}$, \\Hongtao Lu$^{1}$, Lei Zhang$^{2,4}$}

\date{$^{1}$Shanghai Jiao Tong University, $^{2}$OPPO Research Institute \\ $^{3}$Innopeak Technology Inc, $^{4}$The Hong Kong Polytechnic University}

\begin{document}\sloppy

\def\x{{\mathbf x}}
\def\L{{\cal L}}



\maketitle

\begin{abstract}
The Segment Anything Model (SAM) has exhibited outstanding performance in various image segmentation tasks. Despite being trained with over a billion masks, SAM faces challenges in mask prediction quality in numerous scenarios, especially in real-world contexts. In this paper, we introduce a novel prompt-driven adapter into SAM, namely Prompt Adapter Segment Anything Model (PA-SAM), aiming to enhance the segmentation mask quality of the original SAM. By exclusively training the prompt adapter, PA-SAM extracts detailed information from images and optimizes the mask decoder feature at both sparse and dense prompt levels, improving the segmentation performance of SAM to produce high-quality masks. Experimental results demonstrate that our PA-SAM outperforms other SAM-based methods in high-quality, zero-shot, and open-set segmentation. We're making the source code and models available at \url{https://github.com/xzz2/pa-sam}.
\end{abstract}


\section{Introduction}
\label{sec:intro}

Image segmentation stands as a prominent computer vision task with broad-ranging applications, including image editing~\cite{edit}, medical imaging~\cite{medical}, and autonomous driving~\cite{autonomous}. High-quality segmentation goes beyond basic segmentation by providing more detailed masks, particularly for high-resolution images. It not only enables more precise positioning and recognition in perception but also contributes to a deeper understanding of image analysis. Furthermore, high-quality segmentation information can significantly enhance tasks sensitive to details, such as super-resolution~\cite{super}, matting~\cite{matting}, dehazing~\cite{dehazing}, and so on.

The recently developed foundational segmentation model SAM~\cite{SAM} is capable of generating multiple accurate and reasonable masks for arbitrary images based on prompts, showcasing substantial influence and potential advancements in segmentation tasks. Subsequent researches have extended SAM's applications across diverse areas~\cite{one-shot-1,panoptic,medical}. 
Nevertheless, practical applications have revealed the limitation of SAM in high-quality segmentation performance, notably characterized by coarse mask boundaries for objects like tennis rackets and chairs, as well as erroneous predictions for details such as kite strings and insect antennae~\cite{hqsam}. 

\begin{figure}[t]
  \centering
  \includegraphics[width=0.7\textwidth]{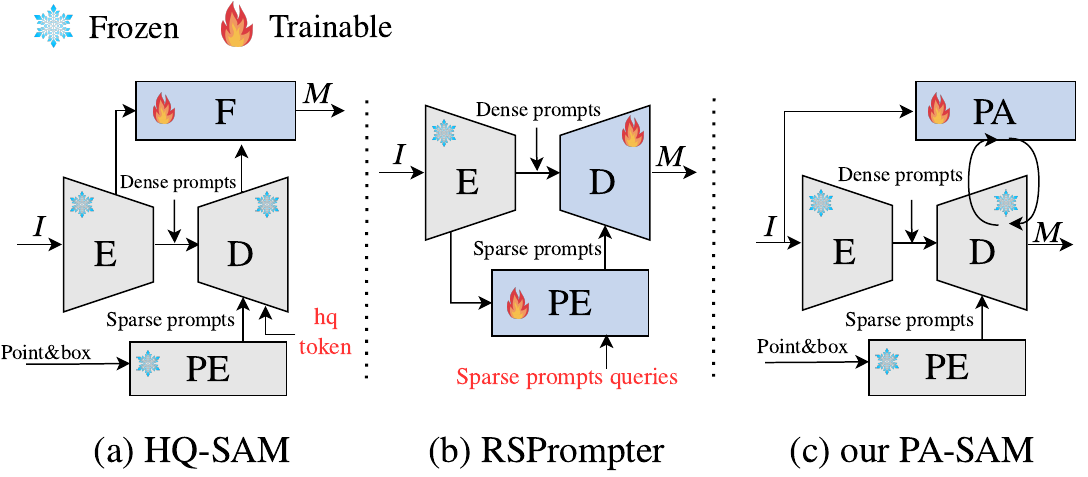}
  \caption{Comparison of different model architectures. `E' means Image Encoder, `D' means Mask Decoder, `PE' means Prompt Encoder, `F' means Feature Fusion Block, and `PA' means Prompt Adapter.}
  \label{fig:model_comparision}
\end{figure}

To address the above mentioned issues, HQ-SAM~\cite{hqsam} introduces a high-quality token to capture more details in the image (see Fig.~\ref{fig:model_comparision}(a)), largely improving SAM's segmentation quality by adding only a few parameters. However, the implicit learning approach used in HQ-SAM makes it challenging to improve SAM's segmentation capabilities, as it primarily focuses on extracting SAM's mask decoder feature for segmentation training, which is isolated from SAM's overall framework. Some prompt-query-based methods~\cite{rsprompter}~\cite{surgical-sam} utilize image features to generate fixed sparse prompts  (see Fig.~\ref{fig:model_comparision}(b)), which can effectively obtain the location of the target object, but they are difficult to capture the detailed object information. Additionally, ensemble~\cite{box1} or augmentation~\cite{augmentation} methods reuse the original input sparse prompts, yielding limited gains in challenging areas.

Therefore, it is highly desired to develop a network that can directly provide SAM with detailed information and improve the mask decoder feature. Intuitively, the most straightforward approach to this end is to provide more detailed annotations, such as additional points or more precise masks. Inspired by this naive intuition, we are wondering if the model could autonomously extract and convey the details to the SAM, thereby significantly improving SAM's segmentation quality without additional user input.

We make such an attempt in this paper. In specific, we introduce the Prompt Adapter Segment Anything Model (PA-SAM), a network designed to investigate uncertain areas in images and incorporate low-level detail information into both dense and sparse prompts to enhance SAM's learning ability for details (see Fig.~\ref{fig:model_comparision}(c)). To capture the details, we propose a prompt-driven adapter to perform adaptive detail enhancement and hard point mining. Unlike conventional adapter~\cite{adapter}, the prompt adapter does not optimize image features, but optimize prompt features to extract detailed information about the network focal area \textbf{(contribution 1)}. We transform the process of mask refinement into the learning of a refined token and an uncertain token, enabling the model to be more sensitive to image details in challenging areas \textbf{(contribution 2)}. Additionally, we propose a hard point mining method based on Gumbel top-k operation, providing direct detailed guidance to the model \textbf{(contribution 3)}. During training, PA-SAM freezes the SAM component and only trains the prompt adapter, thereby preserving the powerful object localization capability of the original SAM while generating high-quality segmentation maps. Our method achieves leading performance on the high-quality dataset HQSeg-44K, with an improvement of $1.7\%$ in mIoU and $2.7\%$ in BmIoU over the previous state-of-the-art. It also demonstrates promising results on zero-shot segmentation and open-set segmentation datasets.

\section{Method}

\subsection{Brief Review of SAM}
The segment anything model (SAM) \cite{SAM}  is a foundational model with powerful zero-shot segmentation capabilities, capable of outputting reasonable masks based on weak annotations. SAM consists of the following components: an image encoder, a mask encoder, a prompt encoder, and a mask decoder. The image encoder transforms input images into $64\times64$ encoded features. The mask encoder encodes masks to dense prompts, while the prompt encoder encodes points or bounding boxes to sparse prompts. The mask decoder, composed of multiple layers of attention, interacts image features with prompt features to output the final segmentation map. Though SAM has demonstrated strong capability in segmentation tasks, its segmentation quality heavily depends on whether the prompts input to the mask decoder can carry detailed information. In the absence of detailed guidance, SAM performs poorly in achieving high-quality segmentation.

\begin{figure}[h]
  \centering
  \includegraphics[width=0.8\textwidth]{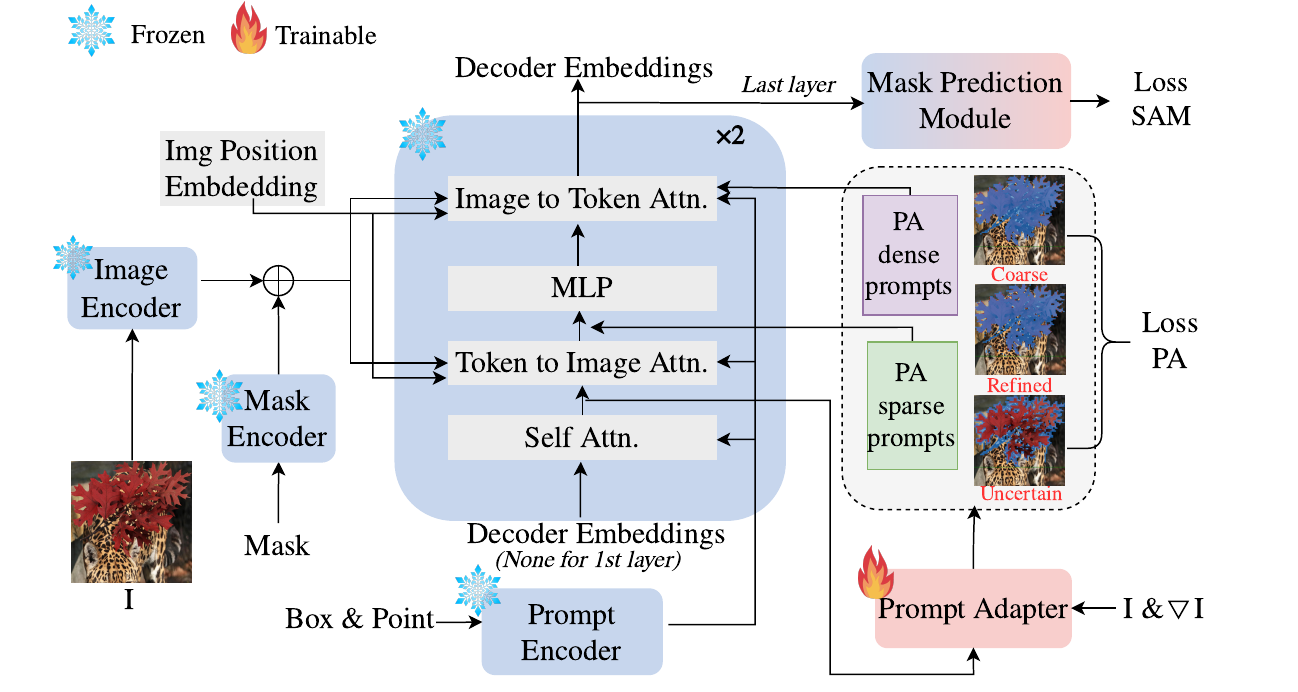}
  \caption{The overall framework of PA-SAM. During the training phase, the SAM parameters are frozen, and only the prompt adapter and the image upsampling module in mask prediction module are trained. 
  During the inference phase, only the output mask from the mask prediction module is used as the final prediction result. The enlarged images of the intermediate masks can be found in Fig. \ref{fig:point}.}
  \label{fig:framework}
\end{figure}

\subsection{Overall Framework of PA-SAM}
To capture high-quality detailed information, our idea is to transform the image details into multi-granularity prompt features and pass them to the mask decoder. That is, we fine-tune SAM in a prompt-driven manner. Based on this idea, we propose a trainable prompt-driven adapter and integrate it into SAM, resulting in our Prompt Adapter SAM (PA-SAM). 

The overall architecture of PA-SAM is shown in Fig.\ref{fig:framework}. PA-SAM combines image features with dense prompts and sends them, along with sparse prompts, to the mask decoder, where the proposed prompt adapter separately transforms the image features and sparse prompts into dense and sparse adapter prompts following each block's self-attention. Subsequently, the output prompt features are reintegrated into PA-SAM in a residual manner to optimize the feature representation of the mask decoder. In our architecture, the model can utilize both detailed and less-detailed information simultaneously, thus improving the quality of segmentation.

\subsection{Prompt Adapter (PA)}
\begin{figure}[t]
  \centering
  \includegraphics[width=0.8\textwidth]{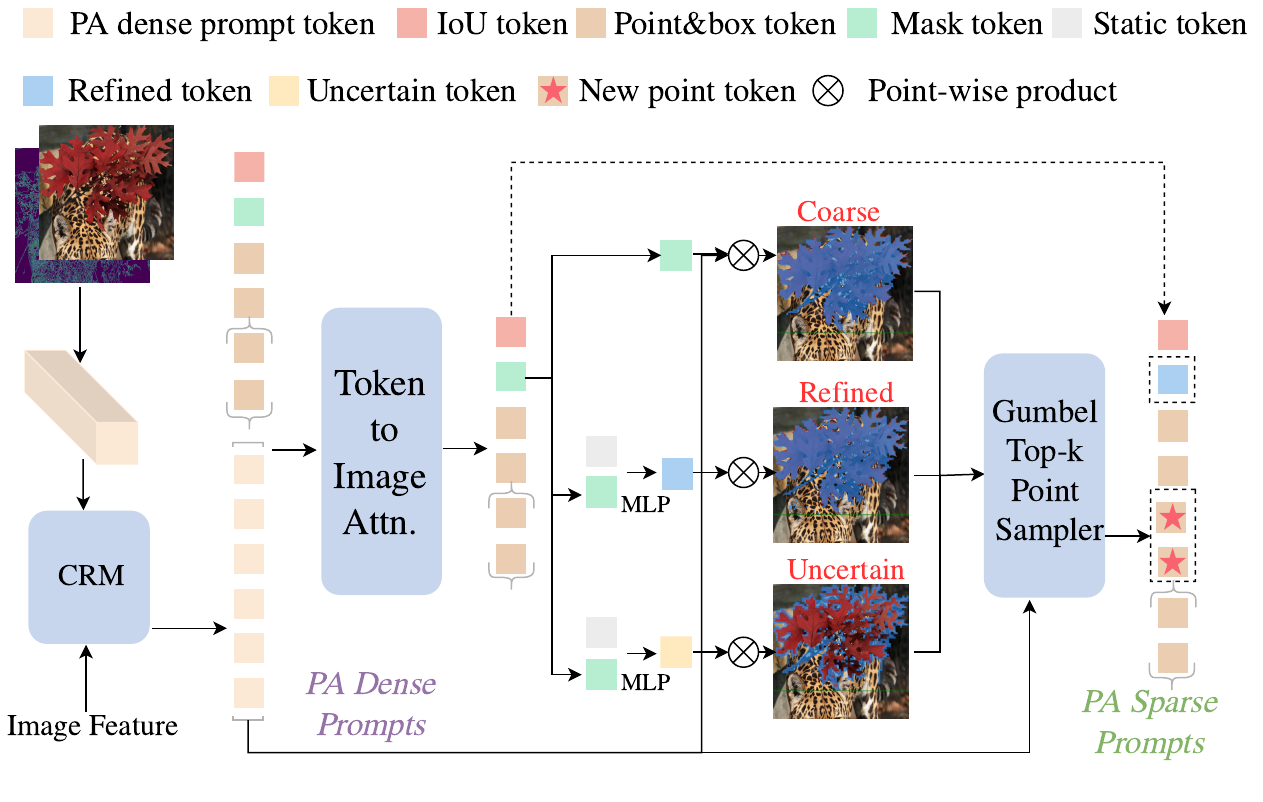}
  \caption{The architecture of the prompt adapter, which achieves adaptive detail enhancement using a consistent representation module (CRM) and token-to-image attention, and implements hard point mining using the Gumbel top-k point sampler.}
  \label{fig:prompt_adapter}
\end{figure}

To improve the network's capability to learn details in uncertain areas, we propose a trainable prompt-driven adapter in the mask decoder of SAM, as depicted in Fig. \ref{fig:prompt_adapter}. This module integrates detailed information into the network through \textbf{Adaptive Detail Enhancement} and \textbf{Hard Point Mining} to adaptively capture relevant detail information based on the original prompts, as detailed below. 

\subsubsection{Adaptive Detail Enhancement}
To capture high-quality detailed information, the prompt adapter performs adaptive detail enhancement to explore detail information from the image and its gradient by \textbf{Dense Prompt Compensation} and \textbf{Sparse Prompt Optimization}.

\textbf{Dense Prompt Compensation.} During image encoding, SAM experiences a significant loss of detailed information due to its $16 \times 16$ downsampling. 
To address this issue, we design a simple compensation module, which encodes the original image $I$ and its gradient $\nabla I$ (such as the Canny operator) as guiding information. Then, by using a consistent representation module (CRM) as cross-attention or guided gate (please refer to the Ablation Study in Section 3.4 for details), it can maintain consistency between output features and image features. In general, the PA dense prompts $x_{pa}$ can be represented by the following formula:
\begin{equation}
    x_{pa} = \texttt{CRM}(W_g[I,\nabla I],x)
\end{equation}
where \texttt{CRM} is the consistent representation module, and $W_g$ denotes convolutional operations.

\textbf{Sparse Prompt Optimization.} We further optimize the sparse prompt features, enabling the flow of detailed information into sparse prompts and enhancing the model's guidance for high-quality image segmentation. Given the original sparse prompts $t_{in}$, we convert them into detailed sparse prompts $t_{pa}$ through a token-to-image cross-attention:
\begin{equation}
    t_{pa} = \texttt{Attention}(q=t_{in},k=x_{pa},v=x_{pa}).
    \label{equ:1}
\end{equation}
This allows us to optimize the sparse prompt representation while retaining the original weakly labeled guidance. 

Additionally, we define the uncertain token $u_{pa}$ to identify challenging areas and refine token $r_{pa}$ to segment them. These tokens are obtained through an MLP after concatenating the mask tokens with their respective static tokens. Then we obtain three different masks: coarse mask $M_C$, refined mask $M_R$, and uncertain mask $M_U$. The intermediate mask $M_{PA}$ for supervising PA-SAM is given as follows:
\begin{equation}
    M_{PA} = M_U \circ M_R + (\mathbf{1}-M_U) \circ M_C.
\end{equation}

\subsubsection{Hard Point Mining}
Based on the adaptive details enhancement, we further propose to incorporate direct guidance for textural details with sparse prompts. To this end, we propose hard point mining, which utilizes $M_C$, $M_R$, and $M_U$ mentioned in sparse prompt optimization to construct guidance for sampling challenging points. Taking positive point sampling as an example, we first construct the initial sampling guidance $\phi^0$. In the training phase, to ensure diversity in the sampling points, we extend the Gumbel-Softmax operation~\cite{gumbel} to the Gumbel top-k operation. In the case of sampling N positive points, the specific process is as follows:
\begin{equation}
\left\{
    \begin{array}{lcc}
        \phi^n = \texttt{flatten}(M_U \circ (M_R-M_C)) & \text { if } & n=0 \\
        \phi^n=\phi^{n-1}+\gamma & \text { elif } & n=1 \\ \phi^n=\phi^{n-1}+ln(1-g^{n-1}) & \text { else }
    \end{array}\right.    
\end{equation}
where $\gamma\sim Gumbel(0,1)$, $n \in N_{sample}$, and $g^n$ represents the Softmax output at the current sample, defined as follows:
\begin{equation}
    g_i^n=\frac{exp(\phi_i^{n}/\tau)}{\sum_{j \in N}exp(\phi_j^{n}/\tau)}.
\end{equation}

All the $g^n$ are summed to obtain $g'$ ($g' = \sum_{n \in N_{sample}}g^n$), which represents the top-k Softmax probability. We then use the straight-through trick~\cite{straight} to obtain the final Gumbel top-k output as follows:
\begin{equation}
    \hat{g} = \texttt{one-hot}(g'_{\texttt{argmax}}) + g'-\texttt{sg}(g'),
\end{equation}
where \texttt{sg} is the stop gradient operator.

We use $\hat{g}$ to perform point sampling on the PA dense prompts $x_{pa}$, which results in $N_{sample}$ positive points. Similarly, the negative point sampling also employs the Gumbel top-k operation, with the initial sampling guided by replacing $\phi^0$ with $\texttt{flatten}(M_U \circ (M_C-M_R))$. This ultimately yields new point prompts $p_{sample}$.  

With sparse prompt optimization and hard point mining, we update the PA sparse prompts $t_{pa}$ as follows:
\begin{equation}
    t_{pa} = [iou_{pa}, r_{pa}, p_{pa}, p_{sample}, b_{pa}],
\end{equation}
where $r_{pa}$ denotes refined tokens and $p_{sample}$ denotes new point prompts.

\section{Experimental Results}
\subsection{Experimental Settings}
\textbf{Implementation Details.} We train PA-SAM using the Adam optimizer with a learning rate of 0.001 and a batch size of 4. The image resolution is $1024\times1024$. We utilize ViT-L as our image encoder backbone and employ BCE loss and Dice loss to supervise both $M_{SAM}$ and $M_{PA}$. The uncertain map's ground truth is obtained by boundary dilation of the ground truth mask. The BCE loss is used for $M_U$.

\begin{table*}[t]
  \centering
  \caption{Results on the high-quality segmentation dataset HQSeg-44K (DIS, COIFT, HRSOD, ThinObject). $\text{SAM}^{*}$ denotes fine-tuning the entire mask decoder of SAM. The results of RSPrompter are replicated on HQSeg-44K. The best and second best results are highlighted in \textcolor{red}{red} and \textcolor{blue}{blue}, respectively.}
  \scalebox{0.85}{
    \begin{tabular}{c|cc|cc|cc|cc|cc}
    \toprule
    \multirow{2}[4]{*}{Model} & \multicolumn{2}{c|}{DIS} & \multicolumn{2}{c|}{COIFT} & \multicolumn{2}{c|}{HRSOD} & \multicolumn{2}{c|}{ThinObject} & \multicolumn{2}{c}{Average} \\
\cmidrule{2-11}          & mIoU  & mBIoU & mIoU  & mBIoU & mIoU  & mBIoU & mIoU  & mBIoU & mIoU  & mBIoU \\
    \hline
    SAM   & 62.0  & 52.8  & 92.1  & 86.5  & 90.2  & 83.1  & 73.6  & 61.8  & 79.5  & 71.1  \\
    $\text{SAM}^{*}$ & \textcolor[rgb]{ 0,  0,  1}{78.9} & 70.3  & 93.9  & 89.3  & 91.8  & 83.4  & 89.4  & 79.0  & 88.5  & 80.5  \\
    HQ-SAM~\cite{hqsam} & 78.6  & \textcolor[rgb]{ 0,  0,  1}{70.4} & 94.8  & 90.1  & \textcolor[rgb]{ 0,  0,  1}{93.6} & \textcolor[rgb]{ 0,  0,  1}{86.9} & 89.5  & 79.9  & 89.1  & \textcolor[rgb]{ 0,  0,  1}{81.8} \\
    RSPrompter~\cite{rsprompter} &77.8 	&69.9 	&94.5 &	88.7 	&92.4 &	86.5 &	90.0& 	79.7 	&88.7 &	81.2   \\
    BOFT-SAM~\cite{boft-sam} & 78.2  & 69.7  & \textcolor[rgb]{ 0,  0,  1}{94.9 } & \textcolor[rgb]{ 0,  0,  1}{90.5} & 93.1  & 86.0  & \textcolor[rgb]{ 0,  0,  1}{91.7} & \textcolor[rgb]{ 0,  0,  1}{80.1} & \textcolor[rgb]{ 0,  0,  1}{89.5} & 81.6  \\
    PA-SAM & \textcolor[rgb]{ 1,  0,  0}{81.5} & \textcolor[rgb]{ 1,  0,  0}{73.9} & \textcolor[rgb]{ 1,  0,  0}{95.8} & \textcolor[rgb]{ 1,  0,  0}{92.1} & \textcolor[rgb]{ 1,  0,  0}{94.6} & \textcolor[rgb]{ 1,  0,  0}{88.0} & \textcolor[rgb]{ 1,  0,  0}{92.7} & \textcolor[rgb]{ 1,  0,  0}{84.0} & \textcolor[rgb]{ 1,  0,  0}{91.2} & \textcolor[rgb]{ 1,  0,  0}{84.5} \\
    \bottomrule
    \end{tabular}}
  \label{tab:high-quality dataset}%
\end{table*}%

\begin{figure}[t]
  \centering
  \begin{subfigure}{0.2\textwidth}
    \includegraphics[width=\linewidth]{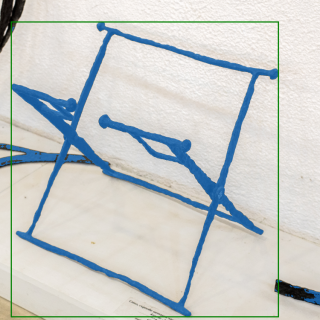}
  \end{subfigure}
  \begin{subfigure}{0.2\textwidth}
    \includegraphics[width=\linewidth]{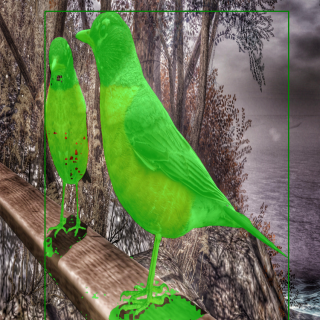}
  \end{subfigure}
  \begin{subfigure}{0.2\textwidth}
    \includegraphics[width=\linewidth]{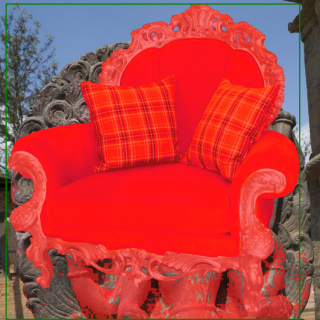}
  \end{subfigure}
  \vskip 0.25\baselineskip
  \begin{subfigure}{0.2\textwidth}
    \includegraphics[width=\linewidth]{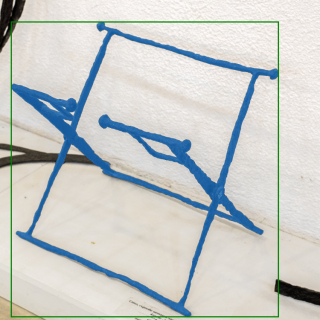}
    \subcaption{}
  \end{subfigure}
  \begin{subfigure}{0.2\textwidth}
    \includegraphics[width=\linewidth]{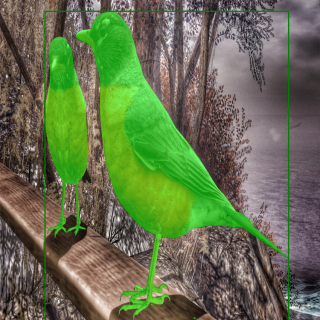}
    \subcaption{}
  \end{subfigure}
  \begin{subfigure}{0.2\textwidth}
    \includegraphics[width=\linewidth]{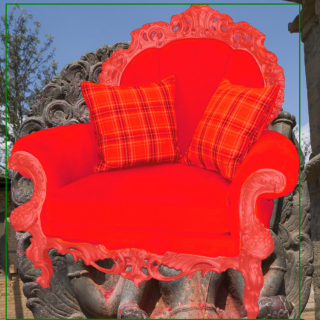}
    \subcaption{}
  \end{subfigure}
  \caption{Visual comparison between HQ-SAM (top row) and PA-SAM (bottom row) on HQSeg-44K.}
  \label{fig:visulization}
\end{figure}

\begin{figure}[t]
  \centering
  \begin{subfigure}{0.3024\textwidth}
    \includegraphics[width=\linewidth]{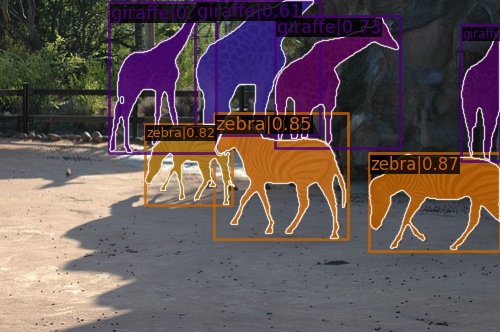}
  \end{subfigure}
  \begin{subfigure}{0.25\textwidth}
    \includegraphics[width=\linewidth]{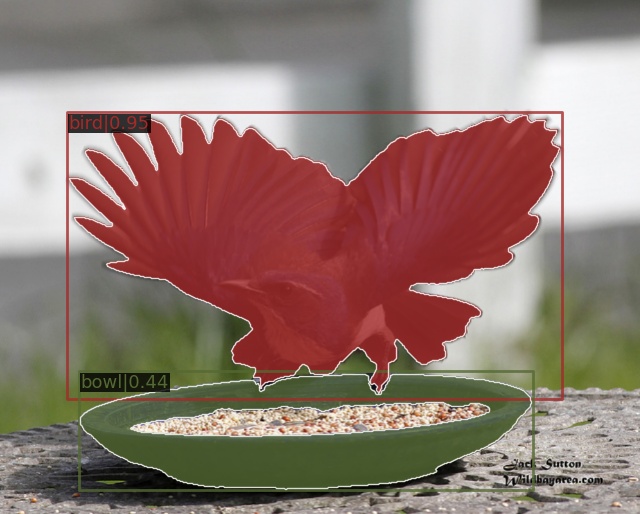}
  \end{subfigure}
  \vskip 0.25\baselineskip
  \begin{subfigure}{0.26\textwidth}
    \includegraphics[width=\linewidth]{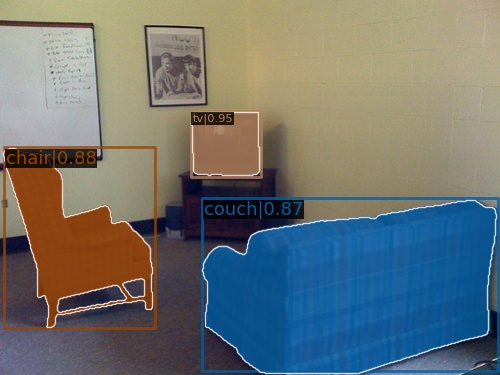}
  \end{subfigure}
  \begin{subfigure}{0.2928\textwidth}
    \includegraphics[width=\linewidth]{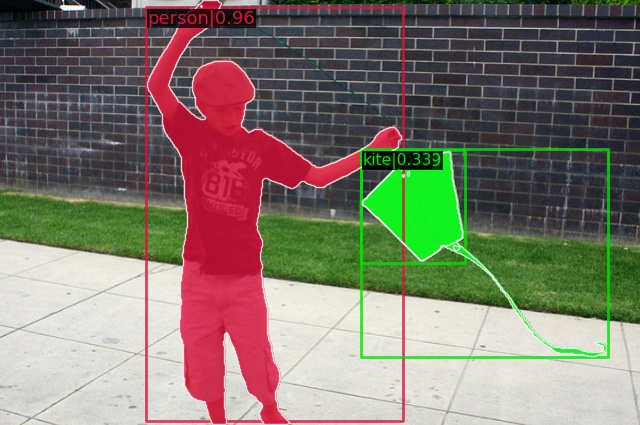}
  \end{subfigure}
  \caption{Zero-shot segmentation results on COCO.}
  \label{fig:coco}
\end{figure}

\textbf{Datasets.} We evaluate our method on HQSeg-44K~\cite{hqsam}, which includes four high-quality segmentation datasets: DIS~\cite{DIS}, ThinObject-5K~\cite{ThinObject}, COIFT~\cite{COIFT} and HR-SOD~\cite{HRSOD}. Furthermore, we evaluate our method on COCO~\cite{COCO} in zero-shot settings. Besides, we experiment on SegInW~\cite{seginw} benchmark (consisting of 25 public zero-shot in-the-wild segmentation datasets) using GroundingDINO~\cite{GroundDION}.

\subsection{High-Quality Segmentation}
Tab. \ref{tab:high-quality dataset} shows the segmentation results on the high-quality segmentation dataset HQSeg-44K. PA-SAM demonstrates significant improvements over HQ-SAM across all the four datasets, with an average increase of 2.1\% in mIoU and 2.7\% in mBIoU. This indicates that optimizing the detailed representation of intermediate features in the mask decoder is more beneficial for generating high-quality segmentation maps than training with final features. By optimizing the features of the mask decoder or fine-tuning the entire mask decoder, the fine-tuning methods such as BOFT-SAM show lower performance (1.7\% lower in mIoU and 2.9\% in mBIoU) than PA-SAM. This is mainly because PA-SAM can perform details learning and exploration for uncertain regions, while other fine-tuning methods basically learn a generic representation that cannot bring significant benefits for high-quality segmentation. In addition, the prompt-query-based methods such as RSPrompter show moderate performance on high-quality segmentation datasets and even perform worse than HQ-SAM. This suggests that the generation of prompts relies not only on the original image information but also on the interaction with the original input prompt. PA-SAM utilizes detailed image information to optimize the representation of the original prompt in the prompt adapter, contributing to its outstanding performance in high-quality segmentation.

Fig. \ref{fig:visulization} presents the visual comparison of PA-SAM with HQ-SAM. When there are objects in the background that closely resemble the target object, PA-SAM can better differentiate them than HQ-SAM. For example, in Fig. \ref{fig:visulization}(a), HQ-SAM incorrectly segments the long stick next to the shelf as part of the shelf, while our PA-SAM accurately segments the components belonging to the shelf. Similarly, in Fig. \ref{fig:visulization}(c), the presence of red paint in the background significantly impacts HQ-SAM, leading to poor segmentation of the chair's bottom. In contrast, our method not only avoids the interference from the red paint but also segments the texture of the chair's bottom more effectively. Furthermore, HQ-SAM exhibits broken masks inherent to SAM (Fig. \ref{fig:visulization}(b), left bird). This is primarily caused by the few or inaccurate sparse prompts, whereas PA-SAM can effectively avoid the occurrence of broken masks through adaptive detail enhancement and hard point mining.

\begin{table}[t]
  \small
  \centering
  \caption{Zero-shot segmentation on COCO. $\text{SAM}^{*}$ fine-tunes the entire mask decoder of SAM. The best and second best results are highlighted in \textcolor{red}{red} and \textcolor{blue}{blue}, respectively.}
    \begin{tabular}{ccccccc}
    \toprule
    Model & AP    & $\text{AP}_{50}$ & $\text{AP}_{75}$ & $\text{AP}_{L}$ & $\text{AP}_{M}$ & $\text{AP}_{S}$ \\
    \midrule
    SAM   & 48.5  & 75.5  & 52.7  & 63.9  & 53.1  & \textcolor[rgb]{ 0,  0,  1}{34.1} \\
    $\text{SAM}^{*}$ & 19.5  &  39.1     &    16.2   & 45.2  & 15.8  & 4.7 \\
    $\text{Adapter~\cite{yang2023aim}}$ & 44.8&	69.5&	48.1&	63.9&	47.8&	29.0  \\
    HQ-SAM~\cite{hqsam} & \textcolor[rgb]{ 0,  0,  1}{49.5} & \textcolor[rgb]{ 0,  0,  1}{75.9} & \textcolor[rgb]{ 0,  0,  1}{53.1} & \textcolor[rgb]{ 0,  0,  1}{66.2} & \textcolor[rgb]{ 0,  0,  1}{53.8} & 33.9 \\
    PA-SAM & \textcolor[rgb]{ 1,  0,  0}{49.9} & \textcolor[rgb]{ 1,  0,  0}{76.1} & \textcolor[rgb]{ 1,  0,  0}{53.9} & \textcolor[rgb]{ 1,  0,  0}{66.7} & \textcolor[rgb]{ 1,  0,  0}{53.9} & \textcolor[rgb]{ 1,  0,  0}{34.5} \\
    \bottomrule
    \end{tabular}%
  \label{tab:coco}%
\end{table}%

\subsection{Zero-Shot and Open-Set Segmentation}
\textbf{Zero-shot segmentation on COCO.} Tab. \ref{tab:coco} reports the zero-shot segmentation results on COCO. We employ the same detector (FocalNet-DINO~\cite{dino}) as HQ-SAM to generate object bounding boxes, which are then used as the sparse prompts to PA-SAM. 
While HQ-SAM brings a 9.6\% mIoU increase over SAM on high-quality segmentation datasets, it only brings 1\% AP increase on COCO. PA-SAM further increases this by 0.4\% AP. Currently, SAM-based zero-shot segmentation methods still exhibit a certain gap in segmentation quality compared to supervised segmentation methods, primarily due to the bottleneck in detection quality of the detector. Detection errors can easily propagate to the segmenter, leading to a decrease in segmentation quality. Compared to HQ-SAM, PA-SAM demonstrates better resistance to detection errors due to the enrichment of sparse prompts through hard point mining, making it more advantageous for zero-shot segmentation. Fig. \ref{fig:coco} illustrates some visual examples on COCO.

\begin{figure}[t]
  \centering
  \includegraphics[width=0.6\textwidth]{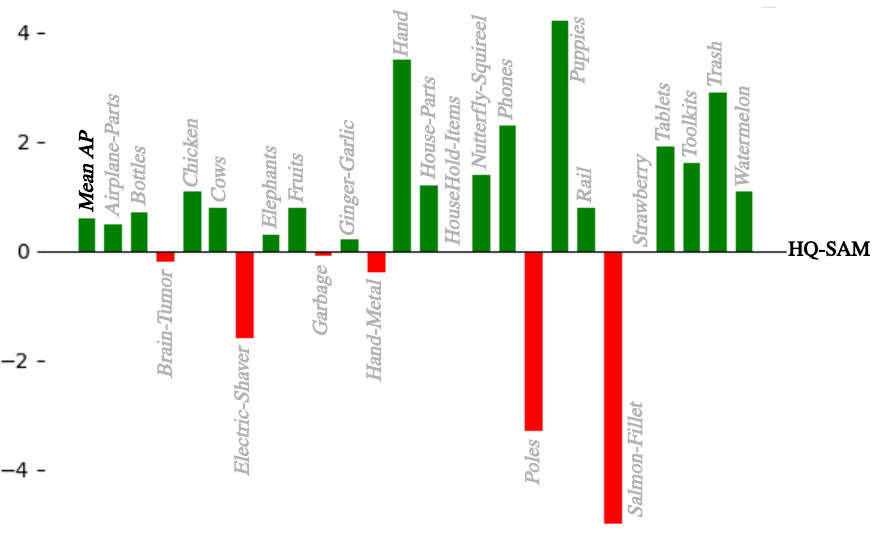}
  \caption{Average precision (AP) of PA-SAM and HQ-SAM across different categories on the Seginw dataset.}
  \label{fig:seginw}
\end{figure}

\textbf{Open-set segmentation on Seginw.} We use Grounding-DINO~\cite{GroundDION} to generate bounding boxes and compare PA-SAM with HQ-SAM by using ViT-H as the backbone. As shown in Fig. \ref{fig:seginw}, PA-SAM achieves an mAP of 50.2\%, which is 0.6\% higher than HQ-SAM. While PA-SAM demonstrates improvement in most categories, its performance is relatively unsatisfactory for a few categories such as electric shaver, butterfly squirrel, and poles. This is mainly caused by the presence of parts in these categories, which significantly differ from the main body. In such cases, PA-SAM may misclassify them into other categories without any prior knowledge.

\begin{table}[t]
  \footnotesize
  \centering
  \caption{Ablation study on DIS dataset. We choose both \textbf{dense and sparse} prompts for adaptive detail enhancement, use \textbf{guided gate} as consistency representation module, set $N_{sample}$ to \textbf{4}, and embed the prompt adapter into the network in a \textbf{parallel} manner to \textbf{one block}.}
  \scalebox{0.8}{
    \begin{tblr}{
  cells = {c},
  cell{1}{1} = {c=3}{},
  cell{2}{1} = {c=3}{},
  cell{3}{1} = {c=3}{},
  cell{4}{1} = {r=12}{},
  cell{4}{2} = {r=2}{},
  cell{6}{2} = {r=2}{},
  cell{8}{2} = {r=4}{},
  cell{12}{2} = {r=3}{},
  cell{15}{2} = {r=2}{},
  vline{4} = {1-4}{},
  vline{2-4} = {4-16}{},
  hline{1,2-4,17} = {-}{},
  hline{6,8,12,15} = {2-5}{},
}
Model      &              &                       & mIoU         & mBIoU        \\
(baseline) &              &                       & 62.0         & 52.8         \\
HQ-SAM     &              &                       & 79.3         & 71.5         \\
PA-SAM      & Enhancement  & Dense                 & 80.7         & 71.6         \\
           &              & \textbf{Dense+Sparse} & 81.0         & 72.2         \\
           & CRM          & \textbf{Guided gate}  & 82.2         & 74.4         \\
           &              & Cross-attention       & \uline{82.8} & \uline{75.5} \\
           & $N_{sample}$ & 0                     & 81.0         & 72.2         \\
           &              & 1                     & 81.7         & 73.5         \\
           &              & \textbf{4}            & 82.2         & 74.4         \\
           &              & 8                     & 82.1         & 73.9         \\
           & Adapter      & Serial                & 79.0         & 71.5         \\
           &              & \textbf{Parallel}     & 82.2         & 74.4         \\
           &              & Fusion                & 80.0         & 73.3         \\
           & Block        & \textbf{One block}    & 82.2         & 74.4         \\
           &              & Two blocks            & 81.8         & 73.9         
\end{tblr}}
  \label{tab:ablation}%
\end{table}%

\subsection{Ablation Study} 

As shown in Tab. \ref{tab:ablation}, we conduct a series of ablation experiments on the adaptive detail enhancement and hard point mining in the prompt adapter, and analyze the embedding method of the prompt adapter. All experimental results are obtained on the high-quality DIS datasets.

\textbf{Adaptive Detail Enhancement.} We compare the results of dense prompt compensation and sparse prompt optimization without using hard point mining. The dense approach can yield results similar to HQ-SAM. When combined with the sparse approach, it leads to a 0.6\% improvement in mBIoU. Regarding the consistency representation module, we observe that replacing the guided gate with cross-attention yields better segmentation results. However, employing image-to-image cross-attention significantly burdens the mask decoder, leading to a substantial reduction in inference speed. Therefore, we opt to utilize the guided gate as the consistency representation module.

\begin{figure}[t]
  \centering
  \begin{subfigure}{0.2\textwidth}
    \includegraphics[width=\linewidth]{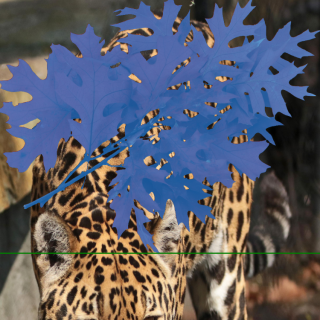}
  \end{subfigure}
  \begin{subfigure}{0.2\textwidth}
    \includegraphics[width=\linewidth]{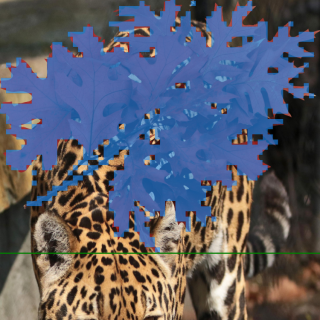}
  \end{subfigure}
  \begin{subfigure}{0.2\textwidth}
    \includegraphics[width=\linewidth]{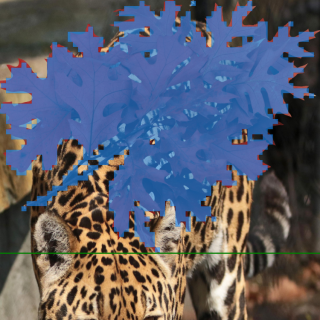}
  \end{subfigure}
  \vskip 0.25\baselineskip
  \begin{subfigure}{0.2\textwidth}
    \includegraphics[width=\linewidth]{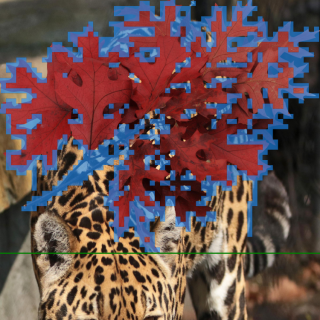}
  \end{subfigure}
  \begin{subfigure}{0.2\textwidth}
    \includegraphics[width=\linewidth]{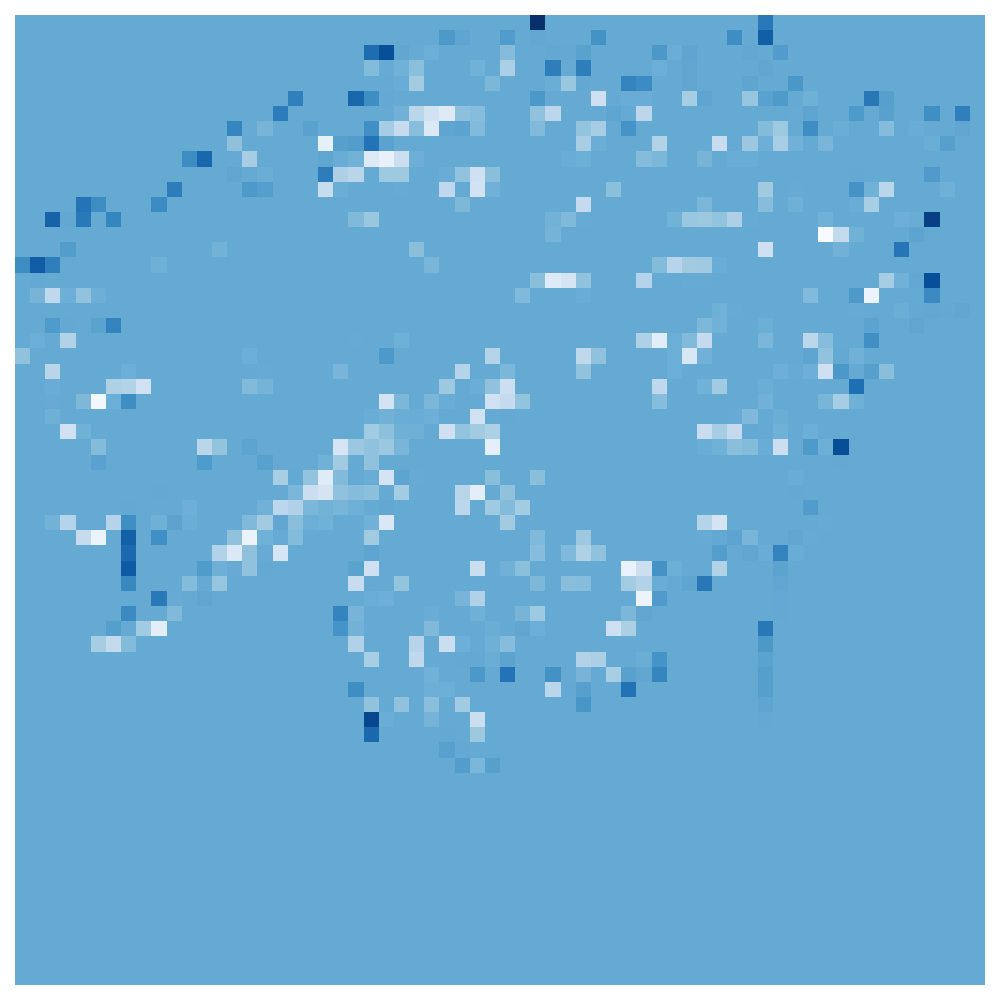}
  \end{subfigure}
  \begin{subfigure}{0.2\textwidth}
    \includegraphics[width=\linewidth]{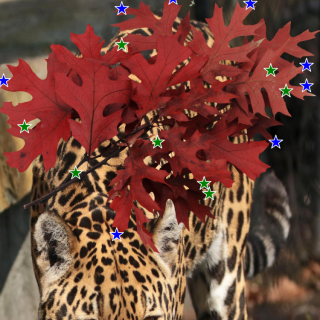}
  \end{subfigure}
  \caption{From left to right and top to bottom: gt mask $M_{GT}$, prompt adapter output mask $M_{PA}$, coarse mask $M_{C}$, uncertain map $M_{U}$, points reference map and sample points (\textcolor{blue}{positive in blue} and \textcolor{green}{negative in green}).}
  \label{fig:point}
\end{figure}

\textbf{Hard Point Mining.} In our ablation study on the number of sampled points $N_{sample}$, we find setting $N_{sample}$ to 4 yields the best performance. A larger $N_{sample}$ leads to decreased performance because the quality of the mask output by the prompt adapter is not very high, and too many sampled points can introduce additional noise. Fig. \ref{fig:point} visualizes the intermediate output of the prompt adapter. $M_{PA}$ can capture many details neglected by $M_C$, but due to the incomplete features and low-resolution output, there is still a certain gap compared to the ground truth. Nevertheless, we observe that the point sampler can effectively capture hard points, such as holes in the leaves (negative points) and the edges of the leaves (positive points), thereby promoting high-quality segmentation by the model.

\textbf{Prompt Adapter Connection.} We compare three methods of embedding prompt adapter and find that the parallel approach yields the best results, as it minimizes the interference with the original mask decoder. Additionally, we add the prompt adapter to both blocks of the mask decoder and add it only to the second block, and observe that adding the prompt adapter to the first block has a negative effect. Upon examining the attention map of the first block, we notice that the mask token in the first block carries very little semantic information, leading to poor mask generation and making it difficult to optimize using the prompt adapter.

\section{Conclusion}
In this paper, we presented a novel prompt-driven adapter for SAM and then developed a high-quality segmentation network, namely PA-SAM, which only needed to fine-tune the prompt adapter. The prompt adapter performed adaptive detail enhancement and hard point mining. It optimized both sparse and dense prompts by mining detailed information from the images. Experimental results demonstrated that PA-SAM improved significantly the segmentation quality without compromising the vanilla SAM's zero-shot segmentation capability, achieving leading performance in high-quality, zero-shot, and open-set segmentation tasks.

\section{References}
\renewcommand{\bibsection}{}
\footnotesize
\bibliographystyle{IEEEbib}
\bibliography{icme2023template}

\begin{thebibliography}{10}

\bibitem{edit}
Tao Yu, Runseng Feng, Ruoyu Feng, Jinming Liu, Xin Jin, Wenjun Zeng, and Zhibo Chen,
\newblock ``Inpaint anything: Segment anything meets image inpainting,''
\newblock {\em arXiv preprint arXiv:2304.06790}, 2023.

\bibitem{medical}
Yichi Zhang and Rushi Jiao,
\newblock ``How segment anything model (sam) boost medical image segmentation?,''
\newblock {\em arXiv preprint arXiv:2305.03678}, 2023.

\bibitem{autonomous}
Xiaoyang Xiao, Yuqian Zhao, Fan Zhang, Biao Luo, Lingli Yu, Baifan Chen, and Chunhua Yang,
\newblock ``Baseg: Boundary aware semantic segmentation for autonomous driving,''
\newblock {\em Neural Networks}, vol. 157, pp. 460--470, 2023.

\bibitem{super}
Zhihe Lu, Zeyu Xiao, Jiawang Bai, Zhiwei Xiong, and Xinchao Wang,
\newblock ``Can sam boost video super-resolution?,''
\newblock {\em arXiv preprint arXiv:2305.06524}, 2023.

\bibitem{matting}
Jingfeng Yao, Xinggang Wang, Lang Ye, and Wenyu Liu,
\newblock ``Matte anything: Interactive natural image matting with segment anything models,''
\newblock {\em arXiv preprint arXiv:2306.04121}, 2023.

\bibitem{dehazing}
Zheyan Jin, Shiqi Chen, Yueting Chen, Zhihai Xu, and Huajun Feng,
\newblock ``Let segment anything help image dehaze,''
\newblock {\em arXiv preprint arXiv:2306.15870}, 2023.

\bibitem{SAM}
Alexander Kirillov, Eric Mintun, Nikhila Ravi, Hanzi Mao, Chloe Rolland, Laura Gustafson, Tete Xiao, Spencer Whitehead, Alexander~C Berg, Wan-Yen Lo, et~al.,
\newblock ``Segment anything,''
\newblock {\em arXiv preprint arXiv:2304.02643}, 2023.

\bibitem{one-shot-1}
Renrui Zhang, Zhengkai Jiang, Ziyu Guo, Shilin Yan, Junting Pan, Hao Dong, Peng Gao, and Hongsheng Li,
\newblock ``Personalize segment anything model with one shot,''
\newblock {\em arXiv preprint arXiv:2305.03048}, 2023.

\bibitem{panoptic}
Xing Lan, Jiayi Lyu, Hanyu Jiang, Kun Dong, Zehai Niu, Yi~Zhang, and Jian Xue,
\newblock ``Foodsam: Any food segmentation,''
\newblock {\em IEEE Transactions on Multimedia}, 2023.

\bibitem{hqsam}
Lei Ke, Mingqiao Ye, Martin Danelljan, Yifan Liu, Yu-Wing Tai, Chi-Keung Tang, and Fisher Yu,
\newblock ``Segment anything in high quality,''
\newblock in {\em NeurIPS}, 2023.

\bibitem{rsprompter}
Keyan Chen, Chenyang Liu, Hao Chen, Haotian Zhang, Wenyuan Li, Zhengxia Zou, and Zhenwei Shi,
\newblock ``Rsprompter: Learning to prompt for remote sensing instance segmentation based on visual foundation model,''
\newblock {\em arXiv preprint arXiv:2306.16269}, 2023.

\bibitem{surgical-sam}
Wenxi Yue, Jing Zhang, Kun Hu, Yong Xia, Jiebo Luo, and Zhiyong Wang,
\newblock ``Surgicalsam: Efficient class promptable surgical instrument segmentation,''
\newblock {\em arXiv preprint arXiv:2308.08746}, 2023.

\bibitem{box1}
Guoyao Deng, Ke~Zou, Kai Ren, Meng Wang, Xuedong Yuan, Sancong Ying, and Huazhu Fu,
\newblock ``Sam-u: Multi-box prompts triggered uncertainty estimation for reliable sam in medical image,''
\newblock {\em arXiv preprint arXiv:2307.04973}, 2023.

\bibitem{augmentation}
Haixing Dai, Chong Ma, Zhengliang Liu, Yiwei Li, Peng Shu, Xiaozheng Wei, Lin Zhao, Zihao Wu, Dajiang Zhu, Wei Liu, et~al.,
\newblock ``Samaug: Point prompt augmentation for segment anything model,''
\newblock {\em arXiv preprint arXiv:2307.01187}, 2023.

\bibitem{adapter}
Neil Houlsby, Andrei Giurgiu, Stanislaw Jastrzebski, Bruna Morrone, Quentin De~Laroussilhe, Andrea Gesmundo, Mona Attariyan, and Sylvain Gelly,
\newblock ``Parameter-efficient transfer learning for nlp,''
\newblock in {\em International Conference on Machine Learning}. PMLR, 2019, pp. 2790--2799.

\bibitem{gumbel}
Eric Jang, Shixiang Gu, and Ben Poole,
\newblock ``Categorical reparameterization with gumbel-softmax,''
\newblock {\em arXiv preprint arXiv:1611.01144}, 2016.

\bibitem{straight}
Aaron Van Den~Oord, Oriol Vinyals, et~al.,
\newblock ``Neural discrete representation learning,''
\newblock {\em Advances in neural information processing systems}, vol. 30, 2017.

\bibitem{boft-sam}
Weiyang Liu, Zeju Qiu, Yao Feng, Yuliang Xiu, Yuxuan Xue, Longhui Yu, Haiwen Feng, Zhen Liu, Juyeon Heo, Songyou Peng, et~al.,
\newblock ``Parameter-efficient orthogonal finetuning via butterfly factorization,''
\newblock {\em arXiv preprint arXiv:2311.06243}, 2023.

\bibitem{DIS}
Xuebin Qin, Hang Dai, Xiaobin Hu, Deng-Ping Fan, Ling Shao, and Luc Van~Gool,
\newblock ``Highly accurate dichotomous image segmentation,''
\newblock in {\em Computer Vision -- ECCV 2022}, Cham, 2022, pp. 38--56, Springer Nature Switzerland.

\bibitem{ThinObject}
Jun~Hao Liew, Scott Cohen, Brian Price, Long Mai, and Jiashi Feng,
\newblock ``Deep interactive thin object selection,''
\newblock in {\em 2021 IEEE Winter Conference on Applications of Computer Vision (WACV)}, Jan 2021.

\bibitem{COIFT}
Jun~Hao Liew, Scott Cohen, Brian Price, Long Mai, and Jiashi Feng,
\newblock ``Deep interactive thin object selection,''
\newblock in {\em 2021 IEEE Winter Conference on Applications of Computer Vision (WACV)}, Jan 2021.

\bibitem{HRSOD}
Yi~Zeng, Pingping Zhang, Zhe Lin, Jianming Zhang, and Huchuan Lu,
\newblock ``Towards high-resolution salient object detection,''
\newblock in {\em 2019 IEEE/CVF International Conference on Computer Vision (ICCV)}, Oct 2019.

\bibitem{COCO}
Tsung-Yi Lin, Michael Maire, Serge Belongie, James Hays, Pietro Perona, Deva Ramanan, Piotr Doll{\'a}r, and C~Lawrence Zitnick,
\newblock ``Microsoft coco: Common objects in context,''
\newblock in {\em Computer Vision--ECCV 2014: 13th European Conference, Zurich, Switzerland, September 6-12, 2014, Proceedings, Part V 13}. Springer, 2014, pp. 740--755.

\bibitem{seginw}
``Segmentation in the wild,'' \url{https://eval.ai/web/challenges/challenge-page/1931/overview?ref=blog.roboflow.com}.

\bibitem{GroundDION}
Shilong Liu, Zhaoyang Zeng, Tianhe Ren, Feng Li, Hao Zhang, Jie Yang, Chunyuan Li, Jianwei Yang, Hang Su, Jun Zhu, et~al.,
\newblock ``Grounding dino: Marrying dino with grounded pre-training for open-set object detection,''
\newblock {\em arXiv preprint arXiv:2303.05499}, 2023.

\bibitem{yang2023aim}
Taojiannan Yang, Yi~Zhu, Yusheng Xie, Aston Zhang, Chen Chen, and Mu~Li,
\newblock ``Aim: Adapting image models for efficient video action recognition,''
\newblock {\em arXiv preprint arXiv:2302.03024}, 2023.

\bibitem{dino}
Hao Zhang, Feng Li, Shilong Liu, Lei Zhang, Hang Su, Jun Zhu, Lionel~M Ni, and Heung-Yeung Shum,
\newblock ``Dino: Detr with improved denoising anchor boxes for end-to-end object detection,''
\newblock {\em arXiv preprint arXiv:2203.03605}, 2022.

\end{thebibliography}

\end{document}